\documentclass[letterpaper, 10 pt, conference]{ieeeconf}  
\usepackage{times}

\IEEEoverridecommandlockouts                              

\usepackage{multicol}
\usepackage[bookmarks=true]{hyperref}
\usepackage{graphicx}
\usepackage{amsfonts}
\usepackage{amsmath,amssymb} 
\usepackage{algorithm}
\usepackage[noend]{algpseudocode}
\usepackage{multirow}
\usepackage{subfig}
\usepackage{caption}
\usepackage{soul} 
\usepackage{bchart}
\usepackage{newfloat}
\usepackage{bbm}
\usepackage{pgfplots}
\pgfplotsset{width=9cm,compat=1.8}

\newcommand{\vect}[1]{\mathbf{#1}} 
\newcommand{\bx}{\mathbf{x}}
\newcommand{\bu}{\mathbf{u}}

\DeclareFloatingEnvironment[
   fileext=loc
]{chart}

\usepackage[colorinlistoftodos,prependcaption,textsize=tiny]{todonotes}

\usepackage[normalem]{ulem}

\begin{document}

\title{High-Speed Robot Navigation \\using Predicted Occupancy Maps}

\author{Kapil D. Katyal$^{1,2}$, Adam Polevoy$^{1}$, Joseph Moore$^{1}$, Craig Knuth$^{1}$, Katie M. Popek$^{1}$
\thanks{$^{1}$Johns Hopkins University Applied Physics Lab, Laurel, MD, USA.
        {\tt\small Kapil.Katyal@jhuapl.edu, Adam.Polevoy@jhuapl.edu}}%
\thanks{$^{2}$Dept. of Comp. Sci., Johns Hopkins University, Baltimore, MD, USA.
        }%
}

\maketitle

\begin{abstract}

Safe and high-speed navigation is a key enabling capability for real world deployment of robotic systems.  A significant limitation of existing approaches is the computational bottleneck associated with explicit mapping and the limited field of view (FOV) of existing sensor technologies.  In this paper, we study algorithmic approaches that allow the robot to predict spaces extending beyond the sensor horizon for robust planning at high speeds.  We accomplish this using a generative neural network trained from real-world data without requiring human annotated labels. Further, we extend our existing control algorithms to support leveraging the predicted spaces to improve collision-free planning and navigation at high speeds.  Our experiments are conducted on a physical robot based on the MIT race car using an RGBD sensor where were able to demonstrate improved performance at 4 m/s compared to a controller not operating on predicted regions of the map.


\end{abstract}


\section{Introduction}

A key objective of mobile robotic systems is to execute safe, reliable motion while avoiding obstacles in the shortest amount of time possible. While mobile robots have demonstrated considerable success in recent years, they still fail to maneuver in environments at the speed and agility of humans. Traditional navigation algorithms require explicit perception, mapping, localization and control for collision free motion.  Often, high-speed navigation using these traditional approaches is severely limited by the sensor's field of view (FOV) as well as the computational requirements needed for explicit mapping.  This requires a robot to frequently reduce its speed to rescan the environment, construct a map and replan a new trajectory.

This is in contrast to human navigation where cognitive psychologists have hypothesized that humans (i) actively differentiate between occupied and free spaces based on observations, (ii) make predictions of occupied spaces beyond line of sight and (iii) use these predictions for navigation to help improve robustness and agility~\cite{doi:10.1177/105971230000800301, buckner2010role,Schwartenbeck411272}. 

\begin{figure}[]
	\centering
    \includegraphics[width=0.70\columnwidth]{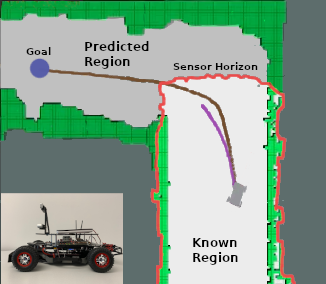}
    \caption{A motivating example of our approach. We leverage occupancy map prediction to allow the robot to plan more robust trajectories at higher speeds compared to those limited by the sensor horizon.}
	\label{fig:first_figure}
\end{figure}

In this paper, we lay the foundation for developing algorithms that provide these capabilities to robotic systems.  Our intuition is that as humans navigate, they leverage spatial cues within the environment to generate predictions of future spaces and use that as part of the planning process. Our objective is to mimic this predictive capability in robotic systems (see Fig.~\ref{fig:first_figure}). Our approach similarly learns to predict spaces beyond the line of sight and further uses the predicted areas as part of the robot controller for planning.


Our specific contributions include:

\begin{itemize}
    \item Novel perception algorithms that predict future occupancy maps using generative neural networks.
    \item A controller that leverages the predicted occupancy map during planning.
    \item Real-world hardware experiments using a robotic car to demonstrate higher speed navigation with improved reliability compared to a controller not operating on predicted regions of the map.
\end{itemize}

\begin{figure*}[]
	\centering
    \includegraphics[width=1.75\columnwidth]{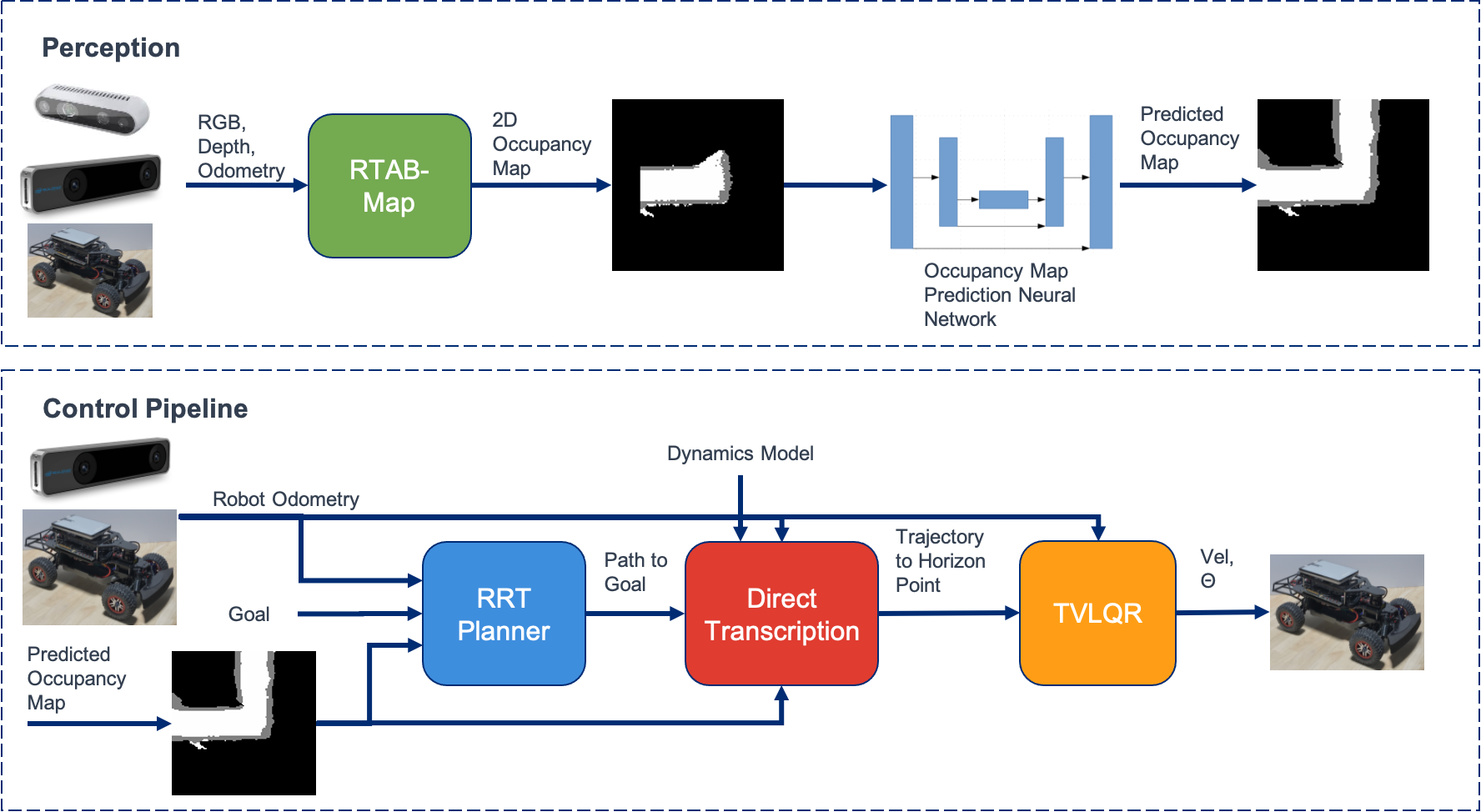}
    \caption{This diagram describes the overall perception and control pipeline.  The perception module receives on-board sensor data from the car and produces a predicted occupancy map using a U-Net style generative neural network.  The control algorithm receives the robot state, predicted occupancy map and goal point and generates collision-free trajectories. }
	\label{fig:hs_sys_diagram}
\end{figure*}

\section{Related Work}

Previous work has presented different strategies to predict unknown parts of the map
.
A variety of methods incorporate map predictions to speed exploration of an environment \cite{chang2007p,shrestha2019learned,Katyal2019,katyal2018,ramakrishnan2020occupancy}. Predominantly, these methods are applied to exploration and therefore do not stress the capability of the map prediction module with respect to the control pipeline. In particular, this paper extends our prior work, \cite{Katyal2019}, by using a balanced classification loss instead of a regression loss in addition to data augmentation and noise suppression techniques required to generate accurate predictions using real data.

An alternative to map prediction is presented in \cite{strom2015predictive}; this method predicts the next best viewpoint for exploration by utilizing experience from previously mapped environments. Our method shows that map prediction effectively increases the amount of information available from the map and improves point-to-point high speed navigation.

Adjacent to map prediction is the problem of planning the shortest path to a goal in an unknown environment. These methods perform some level of inference over the environment to inform motion planning.
For example, in \cite{stein2018learning}, they learn how to plan waypoints to a goal using a partial map of the environment. Unlike our method, this approach does not consider robot dynamics.
Both \cite{richter2014high} and \cite{omri2020} use prior experience to reduce the size of viable environment hypotheses. 
Specifically, \cite{richter2014high} learns the probability of collision of motion primitives whereas \cite{omri2020} utilizes experience-based map predictions in a belief space planner.
Similar to our approach, Elfhafsi \textit{et al.} \cite{Elhafsi2020} incorporate map prediction with global path planning that respects the system dynamics, but all of their verification was done in simulation. 

Existing work has also considered applying reinforcement learning (RL) in tandem with occupancy maps or depth information to navigate an unknown environment.
In \cite{DBLP:journals/corr/KahnVPAL17}, the method uses RGB images and predicts the probability and variance of collision while navigating towards a goal, but is incapable of global planning. 
In \cite{karkus2017qmdp}, the approach models the world as a POMDP and take an end-to-end approach to navigate to a goal.
Additionally the method presented in \cite{li2019deep} uses a partial map as inputs to a deep RL policy to navigate in unknown environments.
While these methods have shown success in application, the data requirements of deep RL policies are large and often impractical in real world scenarios. In addition, our method offers insight in the process by intermediately predicting a map.

A key challenge in high-speed navigation in unknown environments is balancing the speed of navigation against the information gained from its sensors. 
Our method alleviates this tension by learning to infer beyond the FOV of our sensor, but many other methods alter the planning and control pipelines to be reactive to changes in the environment.
For instance \cite{kousik2018bridging, lopez2017aggressive} focus on quickly planning trajectories in order to react to new obstacles.
In \cite{tordesillas2020faster}, the method simultaneously plans two trajectories, a safe trajectory in known space and an aggressive trajectory into unknown space
.
For instance, \cite{gao2019flying} and \cite{Florence2020} plan trajectories and verify safety using only point clouds, similar to \cite{lopez2017aggressive}.
As mentioned earlier, our strategy differs from these methods as they all focus on shortening the time between sensor reading and control generation, whereas our method infers beyond the sensor FOV to amplify the information available.

\section{Preliminaries}

\subsection{Problem Formulation}

Our objective is to enable an unmanned ground vehicle (UGV) to navigate, at a high speed to a goal position in an unknown environment using RGBD and tracking cameras for perception.  Our approach makes use of RGBD mapping, map prediction, and a receding-horizon controller to achieve this objective.


For map prediction, the goal of our network architecture is to learn a function that maps an input occupancy map to an expanded occupancy map that extends beyond the FOV of the sensor.  More formally, we are learning the function
\[
	f : M_{in} \mapsto M_{out}
\]
\noindent where \begin{math}M_{in}\end{math} represents the input occupancy map, and \begin{math}M_{out}\end{math} represents the predicted, expanded output occupancy map.  Components of the function $f$ include an encoder $f_{enc}(M_{in})\mapsto h\in \mathcal{H}$ which maps the input occupancy map to a hidden state and $f_{dec}(h) \mapsto (M_{out})$, which is a decoding function mapping the hidden state to an expanded, predicted occupancy map.

The controller accepts robot odometry state, $\mathbf{x_{robot}}$, the desired goal, $\mathbf{G_{robot}}$, a dynamics model of the robot, $\mathbf{VDM_{robot}}$, and the predicted occupancy map, $M_{out}$ to produce the desired robot velocity, $v$ and turn angle, $\theta$.

\subsection{Platform}\label{sec:platform}

\begin{figure}[]
	\centering
    \includegraphics[width=0.6\columnwidth]{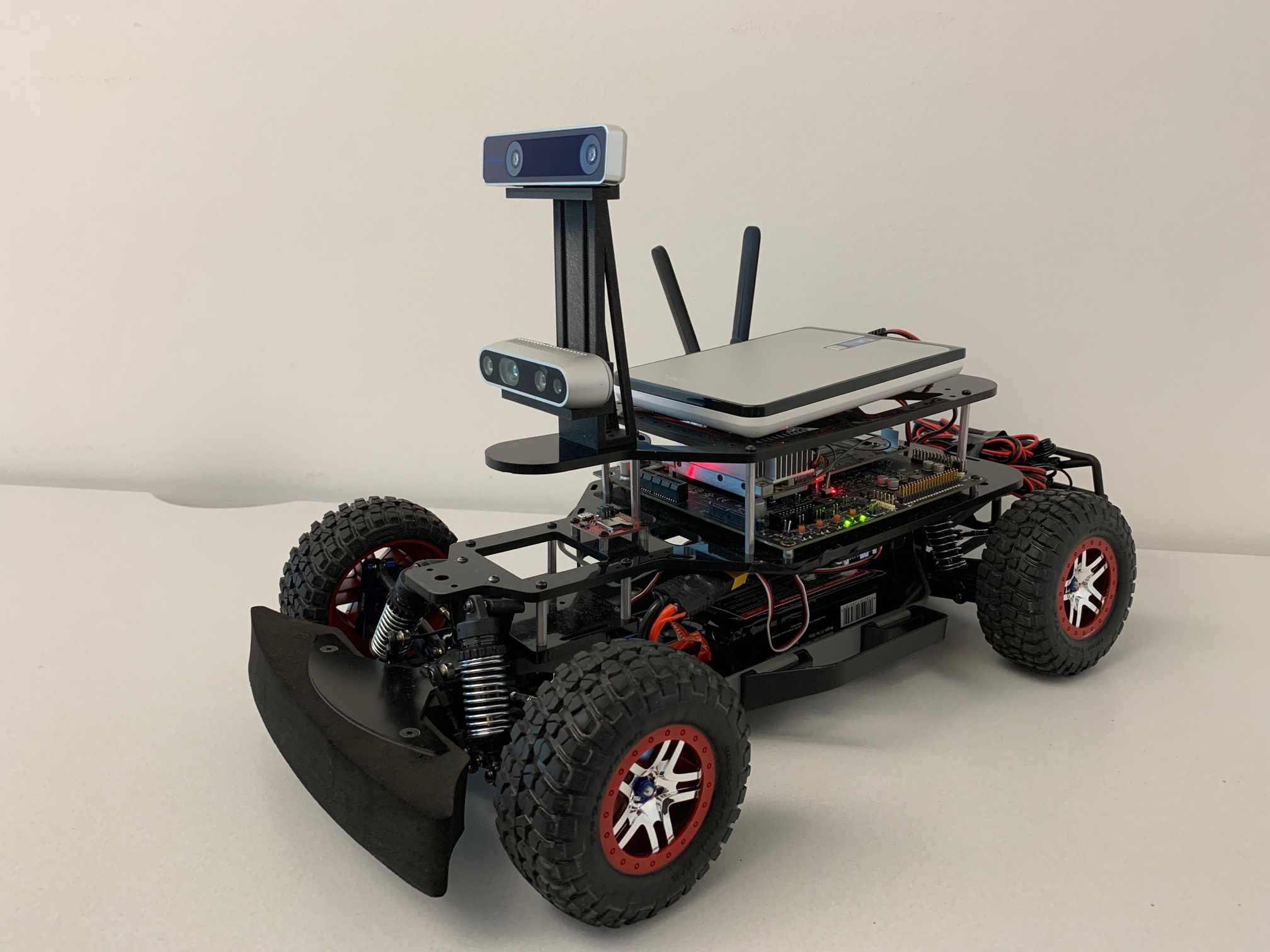}
    \caption{MIT Racecar with Intel Realsense Cameras}
	\label{fig:rc_car}
\end{figure}

The platform we use for our evaluation is the MIT Race car \cite{cain2017high} built on the 1/10-scale Traxxas Rally Car platform, as shown in Fig.~\ref{fig:rc_car}. This RC car has a reported maximum speed of 40 m/s and contains Intel Realsense D435 and T265 cameras.  The onboard Nvidia\textsuperscript{\tiny\textregistered} Jetson TX2 computer runs our perception, mapping and planning software integrated with the Robot Operating System (ROS)~\cite{ros}. Our main interface to the RC car is the variable electronic speed controller (VESC) interface that provides vehicle state information and receives commands including desired velocity and turn angle.

\section{Approach}


\subsection{Perception}

The objective of the perception algorithm is to observe co-registered RGB and depth data to produce an occupancy map for planning. As demonstrated in Fig.~\ref{fig:hs_sys_diagram}, RTAB-Map~\cite{labbe2019rtab} is used to create 2D occupancy maps using the RGB and depth images from a Realsense D435 and visual inertial odometry from a Realsense T265. 

To improve our mapping performance, we limit the D435's depth sensor range to 3 meters, and we apply gradient filtering on the raw depth images.  The depth image gradients are calculated using a Sobel filter with a 5 $\times$ 5 kernel.  All pixels with a gradient magnitude larger than twice the median are discarded.  This removes "ghost noise" near sharp edges in the image. We use RTAB-Map to generate an updated map at approximately 3 Hz while running on the Nvidia\textsuperscript{\tiny\textregistered} Jetson TX2 hardware on the car. 

\subsubsection{Neural Network Architecture}

We use a U-Net style neural network architecture~\cite{DBLP:journals/corr/RonnebergerFB15,pix2pix2017} to receive the occupancy map provided by RTAB-Map and predict an expanded occupancy map. The U-Net neural network is a generative architecture used in several image completion algorithms including~\cite{pix2pix2017,DBLP:conf/iccv/XieLLCZLWD19,DBLP:conf/eccv/YanLLZS18}.  The U-Net network consists of skip connections allowing a direct connection between the layers \begin{math}i\end{math} and \begin{math}n-i\end{math}.  These skip connections enable the option to bypass the bottleneck associated with the downsampling layers and significantly increases the accuracy of predicted occupancy regions~\cite{DBLP:journals/corr/abs-1803-02007}.  Our implementation of the encoder network consists of 7 convolution, batch normalization and ReLU layers where each convolution consists of a $4 \times 4$ filter and stride length of 2.  The number of filters for the 7 layers in the encoder network are: (64, 128, 256, 512, 512, 512, 512).  Similarly, the decoder network consists of 7 upsampling layers with the following number of filters: (512, 1024, 1024, 1024, 512, 256, 128).

\subsubsection{Loss Function}

\begin{figure}[] 
	\centering
    \includegraphics[width=0.45\columnwidth]{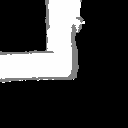}
    \includegraphics[width=0.45\columnwidth]{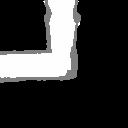}
    \caption{Predicted occupancy map generated without class balancing weight (Left) and with class balancing weight (Right) where white represents unoccupied space, grey is occupied and black is unknown.  The class balancing weight produces stronger edges for obstacles in the predicted occupancy map.}
	\label{fig:weight_balance}
\end{figure}

To train our network, we use a class-balanced cross-entropy loss function as described in~\cite{cui2019classbalancedloss}.  The discrete classes used for labeling each pixel of the occupancy map include occupied, unoccupied and unknown spaces.  Because there are significantly more pixels associated with unoccupied and unknown spaces versus obstacles, we apply class balancing techniques on the cross entropy loss with additional 5$\times$ weight added to the occupied space loss.  This results in predictions where the edges representing obstacles are far more pronounced as seen in Fig.~\ref{fig:weight_balance}.

\subsubsection{Post Processing}

\begin{figure}[] 
	\centering
    \includegraphics[width=0.45\columnwidth]{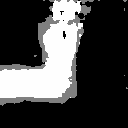}
    \includegraphics[width=0.45\columnwidth]{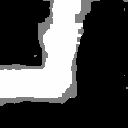}
    \caption{(Left) Generated occupancy map with noise artifacts. (Right) Predicted occupancy map after morphological close operation (white is unoccupied space, grey is occupied and black is unknown).}
	\label{fig:morphological}
\end{figure}

During our testing on the robotic car, we frequently observed small noise artifacts being generated by the neural network, a condition commonly found in generative neural networks~\cite{kaneko2020NR-GAN}. While seemingly minor and transient, these artifacts caused significant instability issues during control as the trajectory planner would often abruptly change the planned path in response to these artificial obstacles or would fail to find a valid trajectory.  To alleviate this, we apply a traditional morphological closing operation with a 5 $\times$ 5 kernel to suppress the noise generated by the neural network with results shown in Fig.~\ref{fig:morphological}. The observed map and the filtered predictive map are combined to create the planning map; any unknown space from the observed map is filled in with data from the predictive map.

\subsubsection{Training Details}

\begin{figure*}[]
	\centering
    \includegraphics[width=1.6\columnwidth]{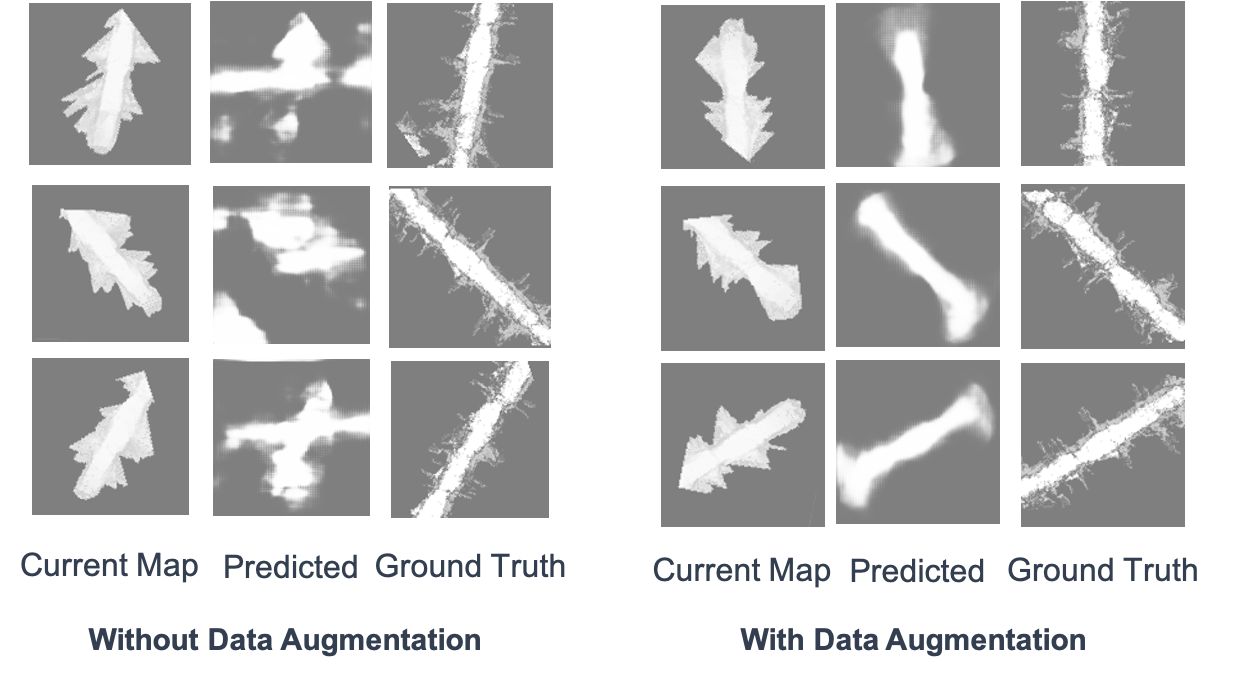}
    \caption{Three examples from our training set of occupancy maps and their resulting expanded predictive map along with the ground truth (white is unoccupied space, light grey is occupied and dark grey unknown).  Augmenting our training data with random rotations, allows the network prediction to be more robust to different environment configurations encountered by the robot.}
	\label{fig:data_augmentation}
\end{figure*}

We generate the datasets in an unsupervised manner. As the robot navigates a new environment, the robot collects data that consists of a submap that corresponds to the current occupancy map based on the sensor's horizon as well as the expanded ground truth map after the environment has been explored. We explored various sizes of the submap and found a map corresponding to 6$\,$m $\times$ 6$\,$m provided by best geometric size of the submap given the characteristics of the sensor.  We used a map resolution of 0.05 meters per pixel so the input occupancy map image resolution was 120 $\times$ 120 pixels.  Further, we experimented with various predicted region sizes and found predicting a region of 7.5$\,$m $\times$ 7.5$\,$m corresponding to an image size of 150 $\times$ 150 provided the optimal accuracy and performance characteristics for the controller.  Further, due to limited amounts of real world data available, we perform data augmentation techniques to apply random rotations to the occupancy map training data. This allows us to be more robust to various hallway configurations as shown in  Fig.~\ref{fig:data_augmentation}.

\begin{figure}[]
	\centering
    \includegraphics[width=0.7\columnwidth]{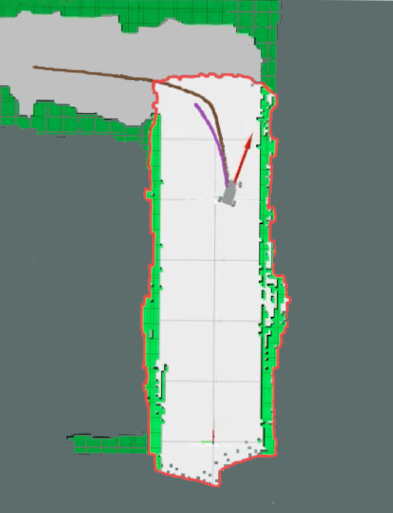}
    \caption{Visualization of system during hardware experiment.  Known map is enclosed by the red boundary.  The brown path is the smoothed RRT path to goal, and the purple path is the local optimized direct transcription trajectory.}
	\label{fig:hardware_experiment}
\end{figure}

\subsection{Control Algorithm}

The objective of the control algorithm as described in Fig.~\ref{fig:hs_sys_diagram} is to compute a collision free path to the goal, generate a series of feasible trajectories to waypoints, and send control commands to the mobile robot to follow the computed trajectory.  We adapt the controller proposed in our prior work,~\cite{basescu2020direct}.  While initially developed for fixed-wing flight, we believe it is particularly well suited for high-speed navigation. The receding horizon allows for rapid replanning while using a dynamically built map, and the trajectory generation and tracking allows for a high-rate, dynamically feasible control output.

\subsubsection{Dynamics Model}

We use a simple bicycle acceleration model to describe the robot's dynamics.  The equations of motion are as follows:

\begin{align}
\dot{x} &= v*cos(\theta) \nonumber \\
\dot{y} &= v*sin(\theta) \nonumber \\
\dot{v} &= u_0 \\
\dot{\theta} &= v * tan(\delta) / L \nonumber \\
\dot{\delta} &= u_1 \nonumber \\ \nonumber
\end{align}

The state is written as $\vect{x}=\begin{bmatrix} x, y, v, \theta, \delta \end{bmatrix}$ where $x$ and $y$ are 2D position, $v$ is forward velocity, $\theta$ is orientation, and $\delta$ is turn angle. The input, $\vect{u}=\begin{bmatrix} u_0, u_1 \end{bmatrix}$, represents acceleration and turn angle velocity, respectively, and $L$ is the wheel base length.

\subsubsection{Control Strategy}

Here, we review the receding horizon controller proposed in \cite{basescu2020direct}, which can be decomposed into three main stages.

In the first stage, a path to goal is generated using a standard rapidly-exploring random tree (RRT) \cite{lavalle1998rapidly}.  The resulting path is pruned and then smoothed using G2 Continuous Cubic B\'ezier Spiral Path Smoothing (G2CBS) \cite{yang2010analytical}. The curvature along the smoothed path, $\begin{bmatrix} x(s), y(s) \end{bmatrix}$, is calculated as: 
\begin{align}
\kappa(s)=\frac{(y''(s)x'(s)-x''(s)y'(s))}{(x'(s)^2+y'(s)^2)^{\frac{3}{2}}} \nonumber
\end{align}
This curvature is then mapped to velocity based upon $v_{max}$ and $v_{min}$, the maximum and minimum velocity. \nonumber
\begin{align}
v(s)=\frac{d\vect{x}}{dt}(s)=v_{max}-\kappa(s)*\frac{v_{max}-v_{min}}{2}
\end{align}
The path's velocity parameterization is used to reparametrize the path by time.
\begin{align}
t=\int_{0}^{s} \frac{1}{v(s)}ds
\end{align}

In our implementation, we modified the RRT to improve performance in a dynamically built map by initializing the RRT tree with the raw RRT path to goal from the previous control iteration.  Before initialization, we check the path for collisions and truncate it if a collision is detected.  This initialization results in faster RRT computation and increased path consistency between iterations.

In the second stage, a dynamically feasible trajectory from the current state to a horizon point is generated.  The horizon point is selected as a time horizon selected along the parameterized RRT path.  We utilize the same direct transcription feasibility problem as formulated in  \cite{basescu2020direct} and refer the reader to their formulation. This approach discretizes the trajectory into $N$ knot points using a variable time interval $dt$. Let $\bx_0(t_k)$ be the position of the robot and $\bu_0(t_k)$, $k < N$, be the input at the $k^{th}$ knot point where $t_{k+1} = t_k + dt$. We modify this feasibility problem by introducing a cost function to penalize large $dt$ (therefore encouraging high speeds) as well as slightly penalizing the input to encourage smoother trajectories. The objective function is shown below.
%
%
\begin{align}
J(\bx_0(t_k), \bu_0(t_k), dt) = \sum_{k=0}^{N-1} \bu_0(t_k)^T \enspace \mathbf{R_c} \enspace \bu_0(t_k) + dt
\end{align}


In order to track the dynamically feasible trajectory, time-varying LQR (TVLQR) is performed in the third stage. The control signal is generated as: 
\begin{align}
\bu(t_k,\bx) &= \vect{K}(t_k)(\bx-\bx_0(t_k))+\bu_{0}(t_k).
\end{align}
where $\vect{K}(t_k)$ is the optimal gain matrix.

\subsubsection{Control Parameters}

The control pipeline executes at a rate of 5$\,$Hz, or a control interval of $T = 0.2 \,$seconds.  The control signal is calculated from the odometry and TVLQR gains at a rate of 50$\,$Hz.  The max time and max iterations for the RRT search were set to 0.05$\,$seconds and 20000 iterations.  The maximum velocity for RRT path parameterization was set to the maximum allowed velocity specified in each hardware trial. For RRT sampling, we use an an obstacle avoidance radius of 0.4$\,$m.

For direct transcription, we used $N = 10$ knot points and a time horizon of $H = 2\,$seconds. We set $\vect{\delta}_f = \begin{bmatrix} 0.1, 0.1, 0.1, 0.25, 100.0 \end{bmatrix}$, $\mathbf{R_c} = diag(0.1, 0.1)$ and use an obstacle radius of 0.35$\,$m.

The costs for TVLQR are as follows:
\begin{align}
Q &= diag(10, 10, 10, 10, 10) \nonumber \\
Q_f &= diag(1, 1, 5, 1, 1) \nonumber \\
R &= diag(1, 1) \nonumber \\ \nonumber
\end{align}

\begin{figure*}
    \centering
    \includegraphics[width=0.48\columnwidth]{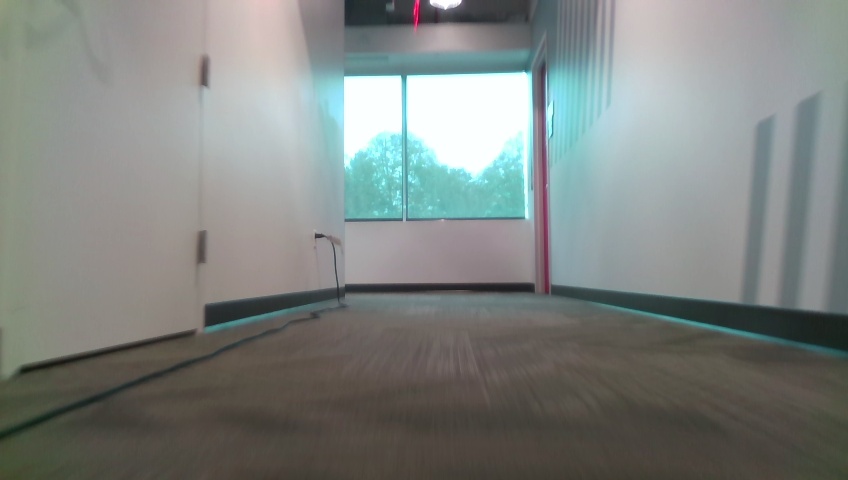}
    \includegraphics[width=0.48\columnwidth]{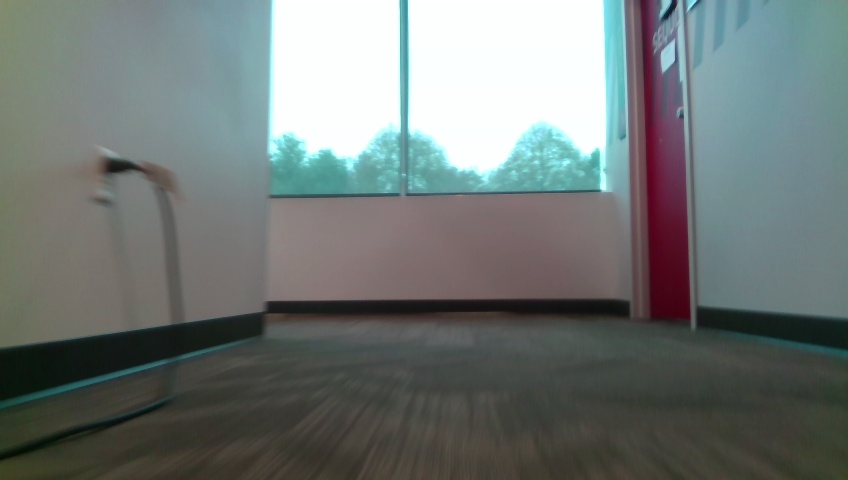} 
    \includegraphics[width=0.48\columnwidth]{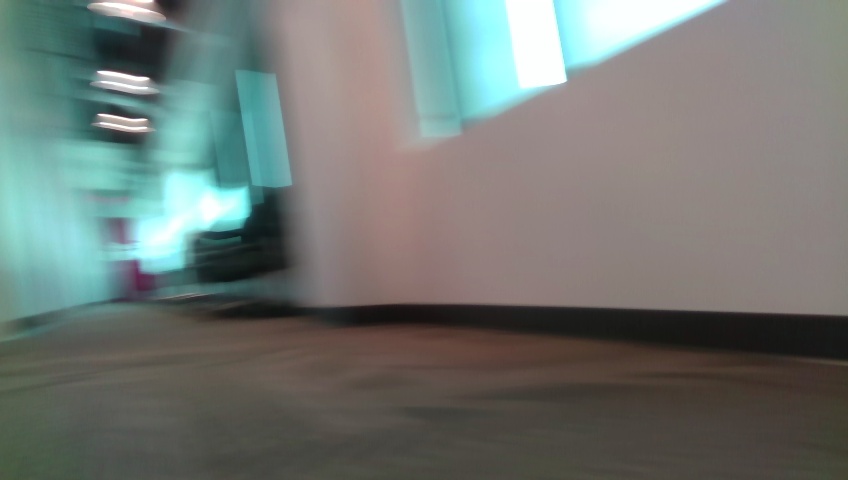}
    \includegraphics[width=0.48\columnwidth]{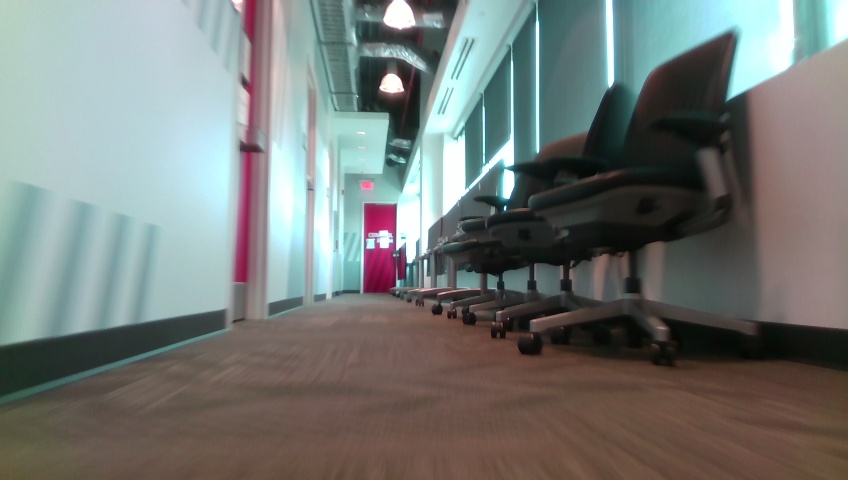}
    \caption{This sequence of images represents a trajectory taken by the robot during our experimental evaluation.}
    \label{fig:car_isc_pics}
\end{figure*}

The acceleration and turn angle control bounds were set to [-2.5, 2.5] m/s and [-1.5, 1.5] rad/s respectively.  The velocity minimum bound was set to 0.5 m/s and turn angle state bounds were set to [-0.3, 0.3] rad.

\section{Experimental Evaluation}

\begin{figure}
	\centering
    \includegraphics[width=1\columnwidth]{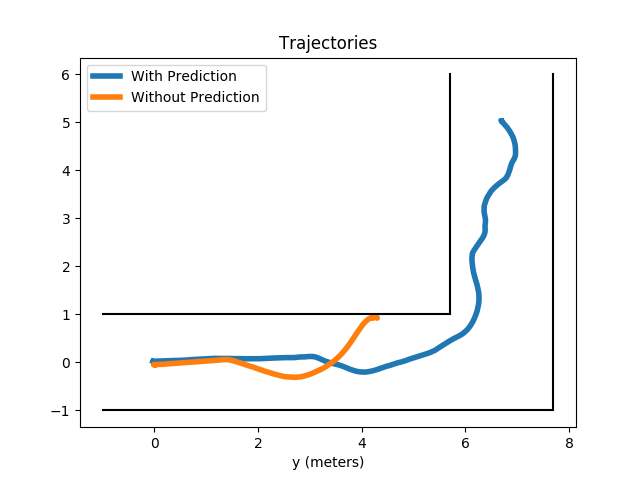}
    \caption{Example trajectories of car with max velocity of 4 m/s with and without map prediction.}
	\label{fig:hardware_trajectories}
\end{figure}

We conduct preliminary hardware experiments to validate our approach using a robotic car based on MIT's open-source race car~\cite{cain2017high} as described in~\ref{sec:platform}.   Our map prediction network was trained on indoor scenes consisting primarily of straight and turning corridors. Similar to~\cite{DBLP:journals/corr/abs-2002-05700}, we conduct both zero-shot and continual learning scenarios. In our zero-shot experiments, we evaluate the performance on new environments (Fig.~\ref{fig:car_isc_pics}) not seen by the robot.  In the continual learning evaluation, we allow the robot to collect data in a semi-supervised manner from the new environment, fine tune the network offline and reevaluate performance.  We assessed the maximum speed allowed by the robot and the number of successful trials, which we defined as reaching the target goal without collision.  A visualization of our system during the hardware experiment is shown in Fig. \ref{fig:hardware_experiment}.  The results are summarized in Table~\ref{table:results}.  Without map prediction, the robot's maximum speed, $v_{max}$, was 3 m/s without collision.  When evaluating without map prediction with the maximum speed of 4 m/s, the robot was only able to successfully reach the goal  1/5 attempts.  With map prediction, we were able to achieve success 3/5 trials.  After allowing the network to fine tune on the new environment, we were able to increase the ratio to 4/5 successful trials showing the ability to continually learn as the robot explores new environments.  For comparison, the trajectories with and without prediction for one of the trials where $v_{max}=4\,$m/s is shown in Fig.~\ref{fig:hardware_trajectories}.  Without map prediction, the robot limited by the sensor's field of view, plans a waypoint in unknown space and is not able to react in time once the map has been updated to reflect the true occupied space.  In contrast, with map prediction, we are able to plan with longer horizons resulting in smoother trajectories that allow the robot to reach the desired goal.

\begin{table}[]
  \centering
   \setlength{\tabcolsep}{1.2mm}
   \renewcommand{\arraystretch}{1.1}
  \begin{tabular}{|c|c|c|}
  \hline
  \textbf{Algorithm} &  \textbf{Max Speed} &  \textbf{Success Rate} \\
  
  \hline
  \hline
  
  Without Map Prediction & 3 m/s & 5/5  \\
  \hline
  Without Map Prediction  & 4 m/s & 1/5  \\
  
  \hline
  With Map Prediction (Zero-shot)  & 4 m/s & 3/5  \\
  \hline
  
 With Map Prediction (Fine Tuned)  & 4 m/s & 4/5  \\
  \hline
  
   \end{tabular}
  \caption{This table captures the results of our preliminary hardware experiments on our modified MIT race car. }
  \label{table:results}
\end{table}

\section{Discussion and Conclusion}


In this paper, we describe an approach that enables high speed navigation based on predicted occupancy maps.  Specifically, we present a generative neural network architecture approach to collect self-supervised training data and training methodology that allows the robot to predict occupied spaces beyond the line of sight of the camera.  Further, we present several engineering solutions in the perception algorithm that were required for prediction to work on real data on the physical hardware including data augmentation, using a class-balanced loss function and traditional computer vision methods for noise suppression.

In addition, we also present a real-time controller that leverages the predicted occupancy map as part of the planning algorithm.  At a max velocity of 3 m/s, 3 Hz mapping rate, 3 m sensor range, and 1 sec time horizon, planned trajectories will almost always be within the known region of the map.  It isn't surprising, thus, that the system is successful without map prediction with these parameters.  When max velocity is increased to 4 m/s, planned trajectories will often be within unknown space of the map (outside of the sensor range).  This can cause trajectories to plan through unseen walls, causing failure.  The predicted occupancy map is able to help address this limitation by providing a longer horizon for planning which accounts for the improved performance at 4 m/s in our preliminary hardware evaluation.

While the results are promising, there are many opportunities for future work.  One area to explore is to extend our preliminary hardware experiments to more thorough training and testing in various indoor and outdoor environments to assess the impact of map prediction for high speed navigation. Another area of future work is to attempt to bypass explicit mapping completely and develop prediction techniques on raw sensor data.  We believe this would improve the performance significantly as map prediction is currently the bottleneck from a computational perspective.  Another area to explore is to improve continual learning techniques as the robot enters new environments. It is unrealistic to expect the training data to capture the full distribution of the environments that the robot will expect to see. We believe further research is needed to improve the data efficiency of continual learning techniques so that the robot can improve performance in real time as it explores new environments.  

In spite of these limitations, we believe we have shown the promise of developing predictive capabilities that improve navigation performance of mobile robots and are continually developing techniques to further extend these capabilities as described above.



\clearpage
\bibliographystyle{IEEEtran}
\bibliography{references}

\end{document}